\documentclass[10pt,twocolumn,letterpaper]{article}

\usepackage{wacv}
\usepackage{times}
\usepackage{epsfig}
\usepackage{graphicx}
\usepackage{amsmath}
\usepackage{amssymb}

\usepackage{balance}
\usepackage{color}
\usepackage{lipsum}
\usepackage{booktabs}
\usepackage[british,american]{babel}
\usepackage{bm,array}
\usepackage[pagebackref=false,breaklinks=true,letterpaper=true,colorlinks,bookmarks=false]{hyperref}
\usepackage{fancyhdr}

\usepackage[font=small,skip=0pt]{subcaption}
\newcommand{\abstrctvspace}{\vspace{-0.2cm}}
\newcommand{\figvspace}{\vspace{-0.4cm}}
\newcommand{\figlblvspace}{\vspace{-0.2cm}}
\newcommand{\tabvspace}{\vspace{-0.0cm}}
\newcommand{\tablblvspace}{\vspace{-0.0cm}}
\newcommand{\reffontsize}{\fontsize{8.8pt}{9.7pt}\selectfont }
\newcolumntype{C}{>{\centering\arraybackslash}p{4.1em}}

\newcommand{\sps}[1]{\mathcal{#1}} \newcommand{\reftab}[1]{\mbox{Table~\ref{#1}}}
\newcommand{\reffig}[1]{\mbox{Figure~\ref{#1}}}
\newcommand{\refsec}[1]{\mbox{Section~\ref{#1}}}
\newcommand{\refeq}[1] {\mbox{Eq.~\ref{#1}}}
\newcommand{\hap}{\mbox{HAT}}
\newcommand{\awa}{\mbox{AwA}}
\newcommand{\apay}{\mbox{aPaY}}
\newcommand{\cub}{\mbox{CUB}}
\newcommand{\genatt}{\mbox{global}~}

\newcommand{\mrkn}[1]{#1}

\fancypagestyle{firstpage}{  \fancyhf{}  \fancyhead[L]{Published as a conference paper at WACV 2015}}

\wacvfinalcopy \def\wacvPaperID{257} 

\ifwacvfinal\fi

\begin{document}

\selectlanguage{american}
\title{How to Transfer? Zero-Shot Object Recognition via\\ Hierarchical Transfer of Semantic Attributes}
\author{Ziad Al-Halah \hspace{2cm} Rainer Stiefelhagen \\
Institute for Anthropomatics and Robotics\\
Karlsruhe Institute of Technology\\
{\tt\small \{ziad.al-halah, rainer.stiefelhagen\}@kit.edu}
}

\maketitle
\ifwacvfinal\thispagestyle{firstpage}\fi

\begin{abstract}
\makeatletter{}Attribute based knowledge transfer has proven very successful in visual object analysis and learning previously unseen classes. 
However, the common approach learns and transfers attributes without taking into consideration the embedded structure between the categories in the source set. 
Such information provides important cues on the intra-attribute variations. 
We propose to capture these variations in a hierarchical model that expands the knowledge source with additional abstraction levels of attributes. 
We also provide a novel transfer approach that can choose the appropriate attributes to be shared with an unseen class. 
We evaluate our approach on three public datasets: aPascal, Animals with Attributes and CUB-200-2011  Birds. 
The experiments demonstrate the effectiveness of our model with significant improvement over state-of-the-art. 
\end{abstract}
\abstrctvspace
\section{Introduction}\label{sec:introduction}
\makeatletter{}Semantic attributes describe the object's  shape, texture and parts. 
They have the unique property of being both machine detectable and human understandable. 
By changing the recognition task from labeling to describing, attributes represent an adequate knowledge that can be easily transferred and shared with new visual concepts. 
Thus, they can be used to recognize unseen classes with no training samples (\ie zero-shot classification). 
\makeatletter{}\begin{figure}[t]
\begin{center}
  \includegraphics[width=0.98\linewidth]{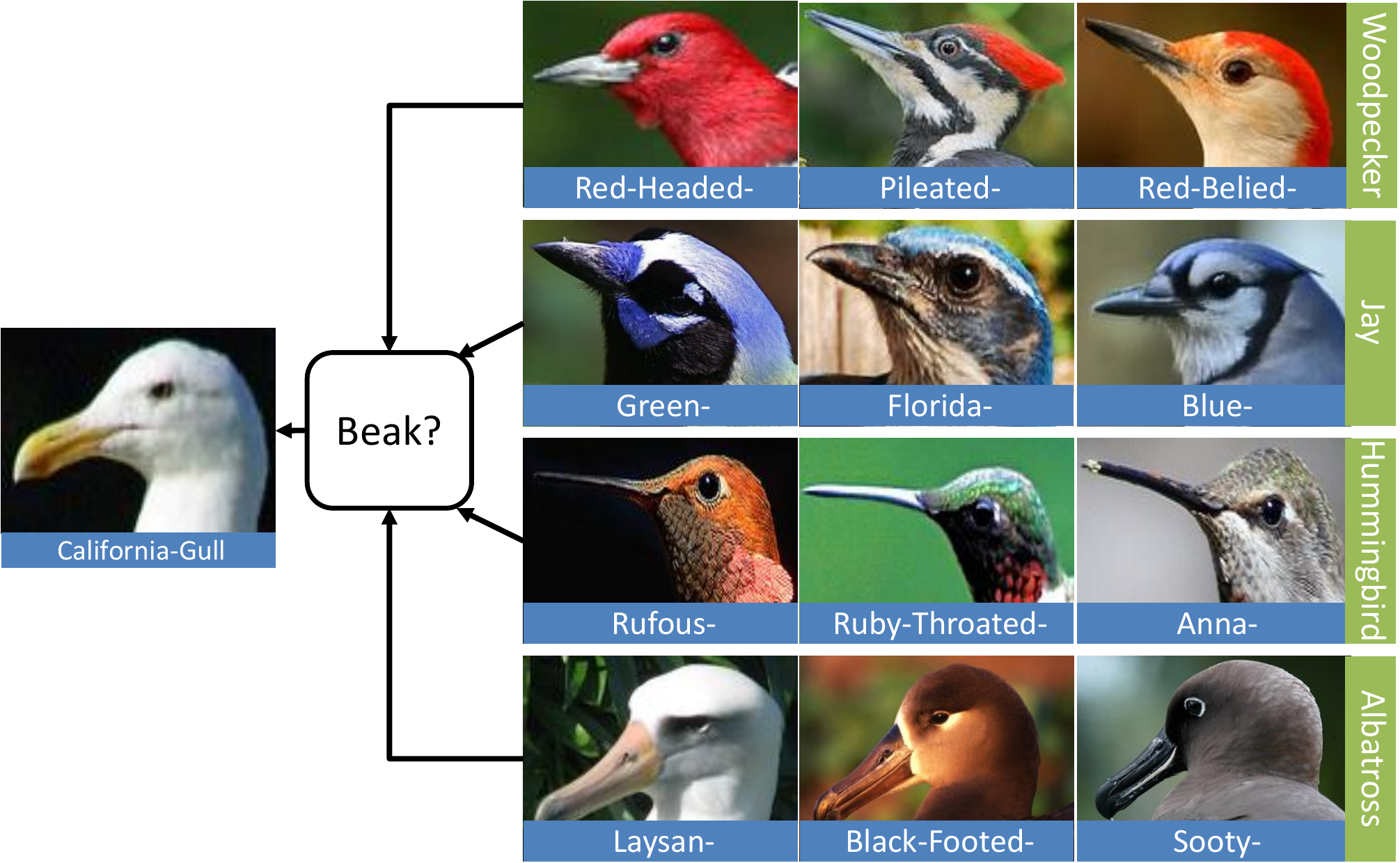}  
\end{center}
\figlblvspace
   \caption{The high intra-attribute variance is better represented at different semantic levels of abstraction. This helps in directing the transfer process to identify the  most suitable source of knowledge to share with a novel class.}
\label{fig:intro}
\figvspace
\end{figure} 

In the prevailing approach, attributes are learned from all seen classes and then reused to describe or classify an unseen one. 
However, this doesn't account for the high intra-attribute variance. 
Using all the seen classes helps in learning visual semantics in a very abstract manner. 
Hence, subsets of classes that share similar attributes cannot be distinguished easily. 
Eventually, the fine properties of the attribute that help in discriminating a group from another are lost when it is learned from all the classes. 
Consider for example the attribute \textit{beak} in \reffig{fig:intro}. 
The \genatt attribute model would learn that a beak is an elongated extension at a certain position relative to the head;  
\ie ignoring the distinctive long thin beak shape of the hummingbird species or the wide curved-end of the albatross species. 
In other words, the \genatt model does not take advantage of the rich information already available in the source dataset. 
This results in transferring less discriminative attributes to the novel class. 
On the other hand, capturing these specific properties of \textit{beak} relative to each subgroup of birds is beneficial. 
It gives us the option to select the most proper type of \textit{beak} to share with the unseen class. 
Accordingly, knowing that both \textit{Gull} and \textit{Albatross} are \textit{Seabirds}, it is intuitive and probably more discriminative to describe the beak of the \textit{California-Gull} as an \textit{albatross-like-beak}. 

In this work we study the benefit of learning attributes at different levels of abstraction, from the most specific that distinguish one class from another to the most general that are learned over all categories. 
We propose a novel hierarchical transfer model that can find the best type of attributes to be shared with an unseen class. 
We evaluate the proposed model on three challenging datasets each with a different granularity of object categories. 
The evaluation shows that significant gain can be achieved with our guided transfer model with  improvements from 26\% and up to 35\% over state-of-the-art in zero-shot classification.

\section{Related work}\label{sec:related_work}
\makeatletter{}Our work relates to two fields in computer vision literature; the attribute-based recognition and hierarchical transfer learning. 

Since the introduction of semantic attributes \cite{Ferrari2008,Farhadi2009,Lampert2009}, they have gained increasing attention from the computer vision community. 
They represent an intermediate layer of semantics and facilitate various tasks in visual recognition like zero-shot learning \cite{Farhadi2009,Farhadi2010,Lampert2009}, aiding object classification and localization \cite{Wang2010}, relative attribute comparison \cite{Parikh2011} and detection of unfamiliar classes \cite{Wah2013}. However, in the prevailing direction the attributes are learned in a \genatt manner from all classes available in the source \cite{Lampert2009,Parikh2011,Farhadi2009,Liu2011}. 
Such methods cannot cope with the high variations in each of the attributes. 
Some approaches jointly model objects, attributes and their correlations \cite{Wang2010,Liu2011} to handle such variations. 
Nonetheless, these correlations cannot be learned for unseen classes since there is no training data. 

Our approach is related to the work on learning class-specific attributes. 
In \cite{Farhadi2009} a set of attributes are learned per class as an intermediate step for feature selection in order to reduce attribute correlations. 
Yu \etal \cite{Yu2013} propose to learn data-driven attributes at the category-level to better discriminate the classes. 
However, data-driven attributes usually carry no semantic meaning; thus, their approach requires user interaction when performing zero-shot learning. 
In \cite{Zhang2013} the concepts in ImageNet are augmented with a set of semantic and data-driven attributes. 
These are used along with the hierarchy to learn a better similarity metric for content-based image retrieval. 
Correspondingly, we propose to explicitly model the intra-attribute variations at different abstraction levels.  
That is, rather than just using class-specific attributes, we expand the notion of the attribute from the most abstract to the most specific driven by the embedded relations between the categories. 

Additionally, hierarchies represent an attractive structure for knowledge transfer and they have been exploited in various ways: parameter transfer \cite{Shahbaba2007,Salakhutdinov2010}, representation transfer \cite{Al-Halah2014} and  bounding box annotations propagation \cite{Guillaumin2012}. 
Of particular relevance to our work is the joint modeling of hierarchy and attributes.  
In \cite{Akata2013}, a ranking classifier is trained using attributes for label embedding;  
showing that using hierarchical labels along \genatt attributes as side information improved the zero-shot performance. 
In the recent work of \cite{Deng2014}, a hierarchy and exclusion graph is learned over the various object categories. 
The graph models binary relations among the classes like mutual exclusion and overlap. 
They also model similar relations between objects and \genatt attributes and use it for zero-shot  classification. 

We exploit the hierarchical structure of the categories in two aspects. 
We leverage the hierarchy to automatically propagate annotations and learn attributes at different levels of abstraction. 
Then we use it in guiding the transfer process to select the most promising knowledge source of attributes to share with novel classes. 
To the best of our knowledge, this has not been done before. 
 
\makeatletter{}\begin{figure*}[!t]
\centering
\includegraphics[width=\linewidth]{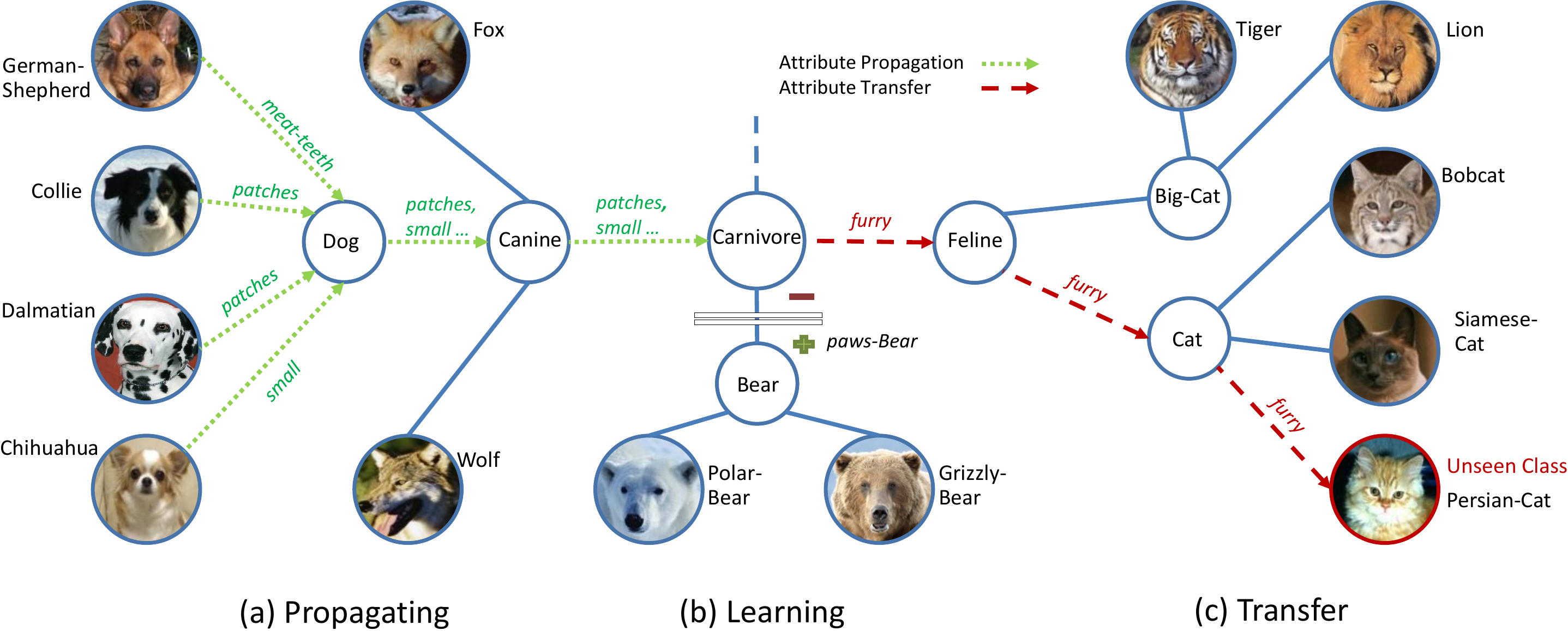}
\figlblvspace
\caption{Illustrative figure of our hierarchical transfer model. We leverage the hierarchy in many aspects: (a) attribute label propagation, (b) modeling the intra-attribute variations and (c) guiding the transfer process to select and share the most relevant knowledge source for a novel class. (Best viewed in color)}
\label{fig:hier}
\end{figure*}

\section{Approach}\label{sec:approach}
\makeatletter{}
Hierarchical representation of concepts and objects is part of the human understanding of the surrounding world. 
We usually try to combine objects into certain groups based on a common criteria like functionality or visual similarity. 
This helps us to better learn the commonality as well as the differences in and across groups. 
The key idea of our approach is to take advantage of the embedded structure in the object category space and extend the notion of \genatt attributes to include different levels of abstraction. 
The object hierarchy groups the classes based on their overall visual similarity; thus provides a natural way to guide the transfer process to share information from the knowledge sources that will most likely contain relative information. 

In the following, we describe the three main steps of our method. 
Starting with a set of classes, \genatt attributes and a hierarchy in the category space: (1) we automatically populate the hierarchy with additional attributes; (2) learn these attributes to capture subtle differences between similar categories; and finally, (3) use a hierarchy-guided transfer to select the proper attributes to share with a novel class. 
We start by defining the notation used throughout the paper.

\textbf{Notation.} 
\makeatletter{}Let $\sps{C}=\{c_k\}^{K}_{k=1}$ be a set of categories, where a subset of these categories $\sps{Q}=\{q_t\}^{T}_{t=1}$ have training samples  while the rest $\sps{Z}=\{z_l\}^{L}_{l=1}$ have none, and $\sps{C}=\sps{Q} \cup \sps{Z}$. 
A set of semantic attributes $\sps{A}=\{a_m\}^{M}_{m=1}$ describe all classes in $\sps{C}$. 
A directed acyclic graph $\sps{H}=(\sps{N},\sps{E})$ defines a hierarchy over the classes, with nodes  $\sps{N}=\{n_i\}^{I}_{i=1}$ and edges $\sps{E}=\{e_{ij}: n_i,n_j\in \sps{N}\}$. 
An attribute $a_m$ at node $n_i$ is referred to as $a_{m}^{n_i}$.  

\subsection{Populating the hierarchy with attributes}\label{sec:hier_attr} 
\makeatletter{}To get the attribute description of the inner nodes we exploit the hierarchy $\sps{H}$ by transferring the attributes annotation in a bottom-up approach from the seen classes $\sps{Q}$ (leaf nodes) to the root. 
For example, the inner node \emph{Dog} (\reffig{fig:hier}a) has the attributes \emph{patches} leveraged from the child nodes \emph{Collie} and \textit{Dalmatian}, which in turn propagates \textit{patches} along other attributes up to node \textit{Carnivore}. Then, the active attributes of node $n_j$ are defined as 
\begin{equation}\label{eq:attinnernodes}
a_{m}^{n_j}=1\hspace{5pt} \mathrm{if} \hspace{5pt}\exists a_m^{n_i}=1 \hspace{5pt}\mathrm{and}\hspace{5pt} n_i\in \mathrm{child}(n_j), 
\end{equation}
where $\mathrm{child}(n)$ is the set of nodes of the subtree rooted with $n$. 
Consequently, the root node of $\sps{H}$ will be described with all attributes of $\sps{Q}$. 
It's important to note that the attribute label propagation is used to find the active attributes at a certain node and to guide the transfer process afterwards.  
This does not change the underlying ground truth attribute labels of samples whether it is class-based or image-based annotation.  

\subsection{Learning at different levels of abstraction}\label{sec:hier_learn}
\makeatletter{}To learn the various attributes classifiers, we first define the support set of an attribute $a_m$, \ie the set of samples that provide evidence of $a_m$. 
An attribute $a_{m}^{n_j}$ in the hierarchy has the support set $\mathrm{supp}(a_m^{n_j})$.
The set contains samples labeled with the attribute of that class ($\mathrm{lbl}(a_m^{n_j})~:~ n_j \in \sps{Q}$), and additionally the samples of its children which share the same attribute with $n_j$, \ie 
\begin{equation}\label{eq:attsupp}
\mathrm{supp}(a_m^{n_j}) = \bigcup\limits_{{n_i}\in\mathrm{child}({n_j})} \mathrm{supp}(a_m^{n_i}) \cup \mathrm{lbl}(a_m^{n_j}).
\end{equation}

To capture the fine differences that characterize an attribute at node $n$, we use a child-vs-parent learning scheme \cite{Marszalek2007}. 
The attribute $a_m^{n_c}$ is learned with the following positive ($T_P$) and negative ($T_N$) sets
\begin{equation}\label{eq:attpos}
T_P= \mathrm{supp}(a_m^{n_c}) \quad T_N= \mathrm{supp}(a_m^{n_p}) - \mathrm{supp}(a_m^{n_c}),  
\end{equation}
where $n_p$ is the parent node of $n_c$. 
For example, the attribute \emph{paws} of node  \emph{Bear} (\reffig{fig:hier}b) is supported by \textit{paws} samples from the classes \emph{Polar-Bear} and \emph{Grizzly-Bear}. 
Then, $a_{paws}^{Bear}$ is learned to capture the differences against the other \textit{paws} samples from parent \textit{Carnivore}. The root $n_r$ of $\sps{H}$ has no parent; hence $\{a_{m}^{n_r}\}$ are learned in the standard 1-vs-all scheme. 
In other words, the root attributes naturally map to the commonly used \genatt attributes in the literature.  

\subsection{Hierarchical transfer}\label{sec:hier_transfer}
\makeatletter{}Having the attributes learned at different levels of abstraction, we then leverage the hierarchy to guide the knowledge transfer process and find the proper attributes to transfer to novel classes.  

Analyzing the recall and precision properties of the attribute classifiers in $\sps{H}$, we generally notice that the attribute predictions towards the leaf level of $\sps{H}$ have higher precision and lower recall than the ones towards the root of the hierarchy, which are characterized with low precision and high recall. 
This is expected from a learning scheme as the one we use. 
While the lower levels capture the discriminative small differences that distinguish a small group of classes against another regarding an attribute (high precision),  
in the higher levels of $\sps{H}$, the common visual properties of the attribute across a larger set of categories is learned (high recall). 
Similar to \cite{Zweig2007}, we find that combining classifiers with such an opposite recall and precision primacy results in an improved performance when compared to the constituent classifiers. 
Furthermore, the combination of classifiers from different levels in the hierarchy increases the robustness of the final classifier against noise that might be produced by the constituent classifiers. 

Accordingly, for a novel class $z_l$ in $\sps{H}$, we transfer the attributes of its ancestors across the different levels of abstraction (\eg $a_{furry}^{PersianCat}$ in \reffig{fig:hier}c), such that:
\begin{equation}\label{eq:attAvg}
s_{z_l}(a_m^{z_l}|\mathbf{x})=\frac{\sum\limits_{n_i\in \mathrm{anc}(z_l)}[[a_m^{z_l}=a_m^{n_i}]]~s_{n_i}(a_m^{n_i}|\mathbf{x})}{\sum\limits_{n_i\in \mathrm{anc}(z_l)}[[a_m^{z_l}=a_m^{n_i}]]}, 
\end{equation}
where $[[\cdot]]$ is the Iverson bracket (\ie $[[P]]=1$ if condition $P$ is true and 0 otherwise), $s_n(a_m^n|\mathbf{x})$ is the score of the  attribute $a_m$ for node $n$ given sample $\mathbf{x}$, and $\mathrm{anc}(n)$ is the set of ancestor nodes of $n$. 
Once the attributes are transferred to $z_l$, the final prediction score $s(z_l|\mathbf{x})$ of the $z_l$ category can be defined by averaging over the attributes of that class as: 
\begin{equation}\label{eq:attSumScore}
s(z_l|\mathbf{x})=\frac{\sum\limits_{m=1}^{M}[[a_m^{z_l}=1]]~s(a_m^{z_l}|\mathbf{x})}{\sum\limits_{m=1}^M [[a_m^{z_l}=1]]}.
\end{equation}
In the following, we refer to our Hierarchical Attribute Transfer model as \hap{}.

\section{Evaluation}\label{sec:evaluation}
\makeatletter{}We evaluate our model on three datasets. 
Each provides different characteristics regarding the granularity of classes. 
This give us the chance to see how the performance of the proposed \hap ~model varies with regards to the complexity of the embedded knowledge in the source set. 
Next, we present the three datasets, the hierarchy and the features we extract from images to train the different attribute models. 

\textbf{Datasets.} 
\makeatletter{}\textbf{(1)} The aPascal/aYahoo (\apay) \cite{Farhadi2009} contains two subsets. 
The first (aP) uses 12,695 images and 20 categories from PASCAL VOC 2008 \cite{pascal-voc-2008}. 
The second (aY) has 12 disjoint classes and 2,644 images collected from the Yahoo image search engine. 
Per-image labels of 64 binary attributes are provided for both subsets. In the predefined zero-shot split, aP is used for training and aY for testing. 

\textbf{(2)} The Animals with Attributes (\awa) \cite{Lampert2009} consists of 30,475 images of 50 classes of animals. 
They are described with 85 semantic attributes on the class level. 
The authors provide a fixed split for zero-shot experiments. 
They select 40 classes for training and 10 for testing. 

\textbf{(3)} The CUB-200-2011 Birds (\cub) \cite{Wah2011} has 200 bird classes and 11,788 images. 
Each image is labeled with 312 attributes. 
Unlike the previous two, there is no predefined split for this dataset. 
For our experiments, we randomly select 150 classes for training and 50 for testing.  

\textbf{Hierarchy.} 
We learn the object hierarchies using the WordNet ontology. 
By querying the ontology with the category labels we extract a tree that brings the classes into a hierarchical ordering. Subsequently, we prune the hierarchy to remove intermediate nodes that have a single child. 

\textbf{Deep Features.}\footnote{The deep features used in this work are available on our website:  \url{https://cvhci.anthropomatik.kit.edu/~zalhalah/}}
\makeatletter{}Motivated by the impressive success of deep Convolutional Neural Networks (CNN), we encode the images using a CNN-based deep representation to train the attribute classifiers.  
We use the CNN model CNN-M2K provided by \cite{Chatfield2014} which has a structure similar to AlexNet \cite{Krizhevsky2012}. 
We also use the BVLC implementation \cite{Jia2014} of GoogLeNet \cite{Szegedy2014} which has a much deeper architecture. 
We follow the best practice found in \cite{Chatfield2014} to extract the deep representation. 
The image is resized to 256 x 256. 
Then 5 image crops obtained from center and corners with their flipped versions are fed to the CNN. 
The output of the last hidden layer is extracted, sum-pooled and L2-normalized to be used as our deep-features to train the different models. 
For all our attribute classifiers, we use linear SVMs \cite{REF08a} with regularized logistic regression. 
The cost parameter C is estimated using 5-folds cross validation.  

\subsection{Attribute Prediction with Deep Features}
\makeatletter{}\begin{table}[!t]
\centering
\scalebox{1.0}{\begin{tabular}{ l | c  c  c  c }
\toprule
Features 		& aPaP	& aPaY 	& AwA	& CUB\\  
\midrule
Shallow			& 84.12		& 70.91				& 71.16	& 60.78\\ CNN-M2K			& 92.82		& \textbf{80.73}	& 78.64	& 76.03 \\ GoogLeNet		& \mrkn{\textbf{93.63}}		& \mrkn{80.03}		& \mrkn{\textbf{79.47}}		& \mrkn{\textbf{76.59}} \\ \bottomrule
\end{tabular} }
\tablblvspace
\caption{Attribute prediction performance (mean AUC) using deep and shallow representations.} 
\label{tab:attrpred_auc}
\tabvspace
\end{table} 
\makeatletter{}We first evaluate the performance of the deep features that we use in learning attribute classifiers. 
We compare their performance with attributes learned using ``shallow'' features. 
We use the precomputed features provided by \cite{Farhadi2009} and \cite{Lampert2009} as the ``shallow'' representation for aPaY\ and AwA respectively. 
For CUB, we encode the images with Fisher vectors based on SIFT and Color descriptors and use that as the shallow representation. In aPaY, we consider two setups: (i) within-category attribute prediction (aPaP), \ie the evaluation is done on the aPascal testing set; (ii) across-category prediction (aPaY), \ie we evaluate on the disjoint aYahoo set. 
In both cases, the attributes are learned using the aPascal training set. 
For AwA and CUB, we use the zero-shot testing setups defined in the previous section. 

In \reftab{tab:attrpred_auc}, we show the performance of the three representations in terms of mean AUC under ROC of the attribute predictions. 
The deep-feature models constantly outperform thier counterpart across all datasets with an increase between \mrkn{7}\% to \mrkn{15}\%. 
This high performance of the deep features in attribute prediction is expected. 
CNNs automatically learn to capture features with varying complexity at each layer. 
While the lower layers learn features like edges and color patches, the higher layers learn much complex structures of the object like parts \cite{Zeiler2014}. 
Many of these correspond directly to semantic attributes which explains the impressive performance of the deep representation using our simple linear classifiers. 

\subsection{Zero-Shot Classification}
\makeatletter{}\begin{table}[!t]
\centering
\scalebox{0.9}{\begin{tabular}{ l | l | c  c  c }
\toprule
Model 				& Features	& aPaY	& AwA 	& CUB\\  
\midrule
DAP \cite{Lampert2013}  	&Shallow	& 19.1	& 41.4	& - \\
IAP \cite{Lampert2013}  	&Shallow	& 16.9	& 42.2	& - \\
AHLE \cite{Akata2013}   	&Shallow	& -		& 43.5	& [17.0] \footnotemark \\
HEX \cite{Deng2014}	   		&DECAF		& -	 	& 44.2	& - \\
TMV-BLP \cite{Fu2014}   	&Shallow 	& -		& 47.1  & - \\
\hap{} (ours)     			&DECAF		& -		& \mrkn{\textbf{48.9}}	& -  \\ \midrule
DAP      			   		&CNN-M2k	& 31.9	& 54.0	& 33.7  \\
ENS			      	   		&CNN-M2k	& 43.1	& 57.7	& 37.3  \\ \hap{} (ours) \cite{Al-Halah2015} 		&CNN-M2k	& \textbf{46.3} & \textbf{68.8}& \textbf{48.6} \\ \midrule
DAP					   		&GoogLeNet	& \mrkn{35.5}	& \mrkn{59.9}	& \mrkn{36.7}  \\ ENS	  	    				&GoogLeNet	& \mrkn{42.8}	& \mrkn{63.5}	& \mrkn{39.4}  \\ \hap{} (ours)     			&GoogLeNet	& \mrkn{\textbf{45.4}}	& \mrkn{\textbf{74.9}}	& \mrkn{\textbf{51.8}}  \\ \bottomrule

\end{tabular} }
\tablblvspace
\caption{Zero-shot multi-class accuracy on the three datasets.} 
\label{tab:zeroshot_acc}
\tabvspace
\end{table}
 
\makeatletter{}\begin{table}[!t]
\centering
\scalebox{1.0}{\begin{tabular}{ l | l | c  c  c }
\toprule
Model 				& Features	& aPaY  & AwA 	& CUB\\  
\midrule
DAP      			&CNN-M2k	& \textbf{87.3} & 88.5 & 82.2 \\
ENS	      			&CNN-M2k	& 85.6 & 88.5 & 89.5 \\
\hap{} (ours)   	&CNN-M2k	& 87.1 & \textbf{92.0} & \textbf{94.9} \\
\midrule
DAP      			&GoogLeNet		& \mrkn{\textbf{88.0}} & \mrkn{89.8} & \mrkn{83.9} \\
ENS	      			&GoogLeNet		& \mrkn{84.7} & \mrkn{90.1} & \mrkn{90.7} \\
\hap{} (ours)   	&GoogLeNet		& \mrkn{86.7} & \mrkn{\textbf{94.1}} & \mrkn{\textbf{95.5}} \\
\bottomrule
\end{tabular} }
\tablblvspace
\caption{Zero-shot mean AUC under ROC curve for the test classes.} 
\label{tab:zeroshot_auc}
\tabvspace
\end{table} 
\makeatletter{}To populate the hierarchy with attributes (\refsec{sec:hier_attr}), our model requires class-based attribute descriptions. 
Hence, for aPaY and CUB we average all image-based attribute vectors of each class to calculate the class-attribute  occurrence matrices. 
Then, the binary class-attribute notations are created by thresholding the resulting occurrence matrix at its overall mean value. 
Along with our Hierarchical Attribute Transfer model (\hap), we also train two common baselines for \genatt attributes using our deep features:  
(i) The Direct Attribute Prediction model (DAP), where the class prediction is formulated as a MAP estimation \cite{Lampert2009}; 
(ii) The Ensemble model (ENS), that combines the predictions of the attributes using a sum formulation similar to the one we use in \refeq{eq:attSumScore} but based on \genatt attributes \cite{Rohrbach2011}. 
For ENS and \hap{}, we normalize the prediction scores of the classes to have zero mean and unit standard deviation. 

\textbf{\hap{} vs. state-of-the-art.} 
In \reftab{tab:zeroshot_acc}, we report the normalized multi-class accuracy on the three test sets. 
Our model outperforms the state-of-the-art on the three datasets with a wide margin. 
In \cite{Akata2013}, a model that uses object hierarchy as side information is used to learn attributes based on Fisher vectors. 
In the recent work of \cite{Deng2014}, a hierarchical model of objects is learned using deep features similar to the one we use. 
Finally, \cite{Fu2014} a transductive multi-view embedding is learned using global attributes, word space and low-level features. 
Nonetheless, our model improves over the best state-of-the-art results by \mrkn{26}\% (aPaY), \mrkn{28}\% (AwA) and \mrkn{35}\% (CUB). 
\footnotetext{Although we followed a similar split on CUB as in \cite{Akata2013}, the actual classes may differ and the results may not be directly comparable.}

\begin{figure*}[!t]
\begin{flushleft}
{\fontsize{9.0pt}{1em}\selectfont 
\begin{tabular}{C C C C C C C C C C}
chimpanzee&giant-panda&leopard&persian-cat&pig&hippo-potamus&humpback-whale&raccoon&rat&seal\\
\end{tabular}
}
\end{flushleft}
\centering
\begin{subfigure}[b]{1.0\linewidth}
	\centering
	\includegraphics[width=\linewidth]{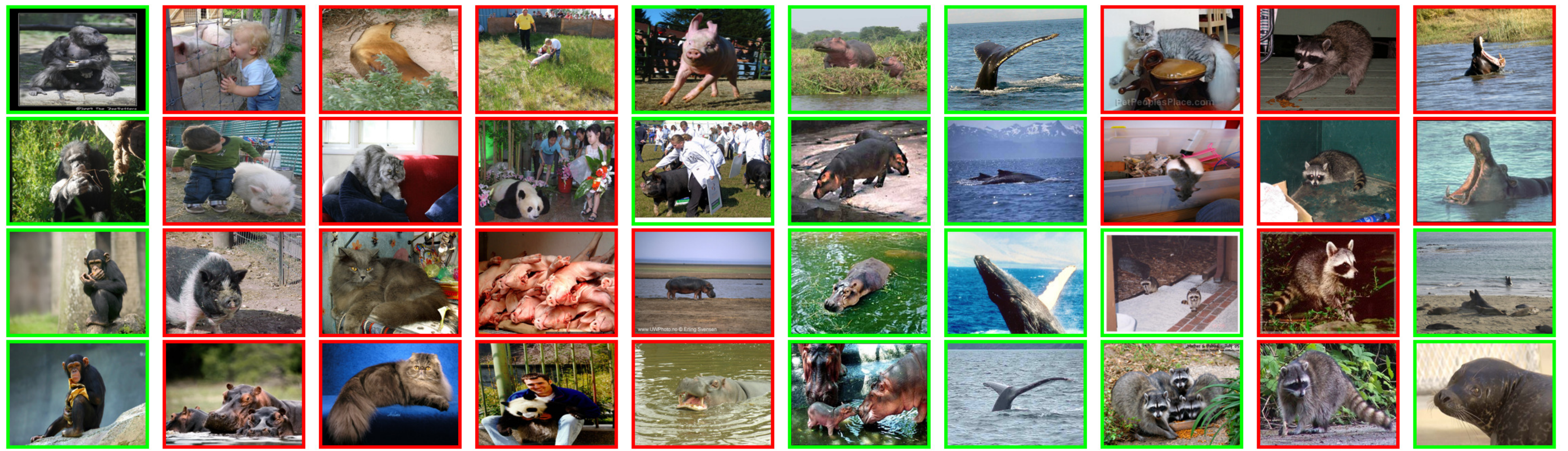}
	\caption{DAP }\label{fig:awa_dap}
\end{subfigure}\\
\begin{subfigure}[b]{1.0\linewidth}
	\centering
	\includegraphics[width=\linewidth]{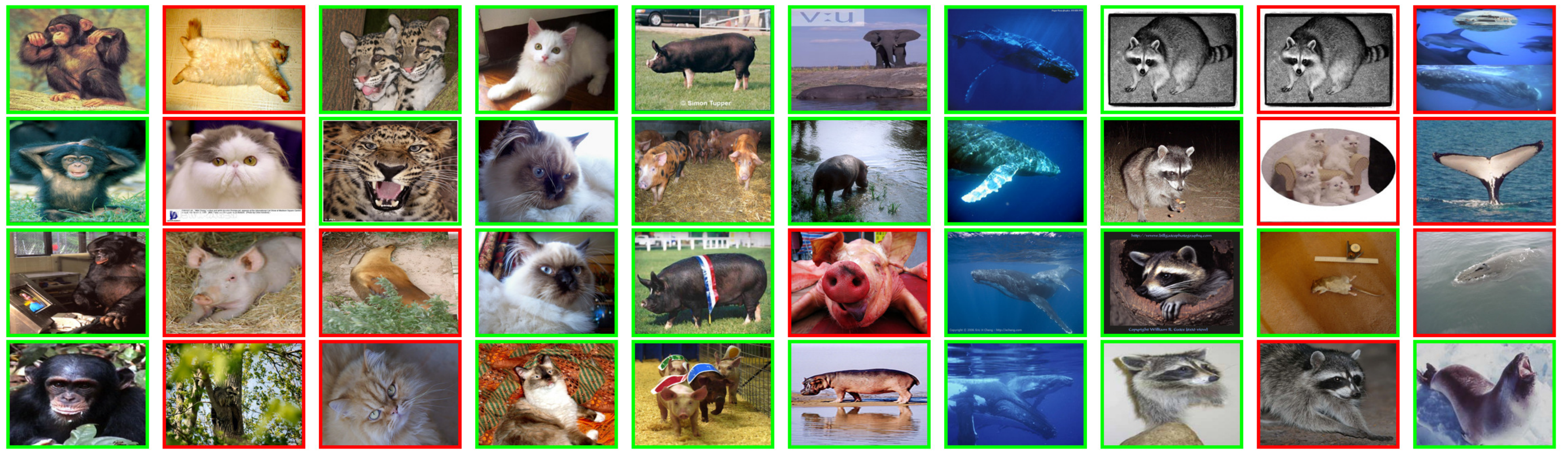}
	\caption{ENS }\label{fig:awa_dap}
\end{subfigure}\\
\begin{subfigure}[b]{1.0\linewidth}
	\centering
	\includegraphics[width=\linewidth]{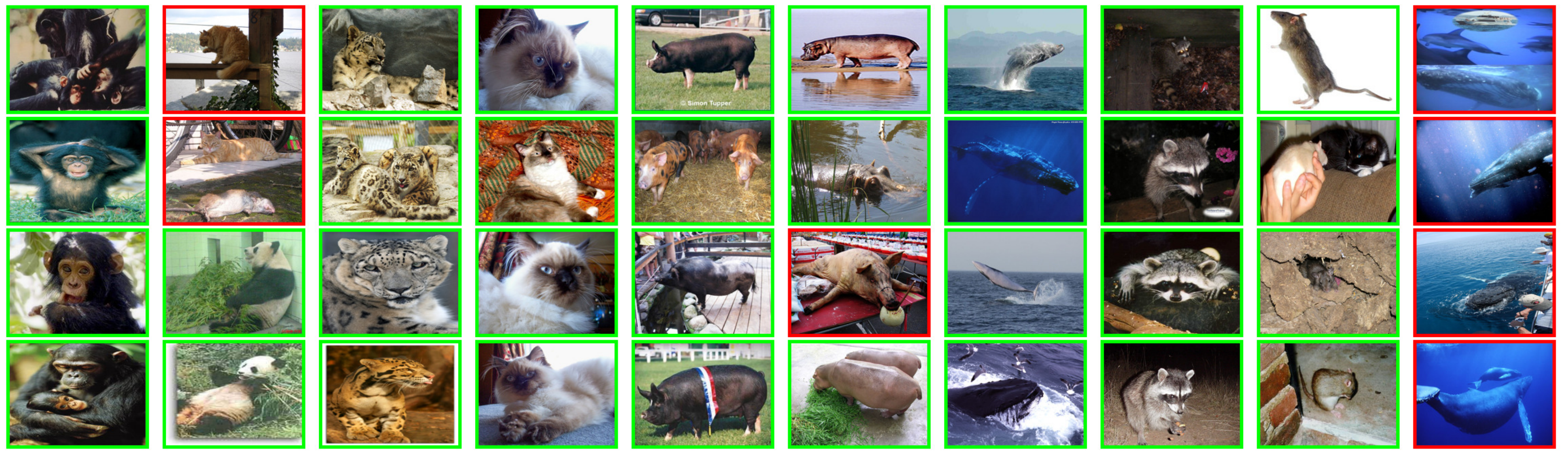}
	\caption{\hap{} (ours)}\label{fig:awa_hat}
\end{subfigure}~~
\figlblvspace
\caption{\mrkn{The highest scoring results of DAP, ENS and \hap{} (CNN-M2K) for each test class in AwA. (Best viewed in color)}}
\label{fig:awa_visres}
\figvspace
\end{figure*}

\textbf{\hap{} vs. deep-feature baseline.} 
Compared to our strong baseline with deep features (DAP and ENS), \hap{} still performs the best in terms of both multi-class accuracy and mean AUC (\reftab{tab:zeroshot_auc}). 
\reffig{fig:awa_visres} shows the highest ranking results obtained by the three models for each test class in the AwA dataset. 
While distinctive classes like \textit{Chimpanzee} and \textit{Humpback-Whale} are correctly classified by all models, both DAP and ENS confuse visually similar classes that share many global attributes like \textit{Leopard} \& \textit{Persian-Cat} and \textit{Rat} \& \textit{Raccoon}, and more samples are wrongly ranked high by these models. To the contrary, \hap{} learns the fine differences among the shared attributes of these classes which helps in discriminating them efficiently. 
For example, \hap{} learns the differences between the attributes of \textit{Big-Cat} and \textit{Cat} (\reffig{fig:hier}c) which facilitates the separation among the novel classes  \textit{Leopard} and \textit{Persian-Cat} (\reffig{fig:awa_visres}). 
Furthermore, we find that normalizing the prediction scores of the novel classes (\refeq{eq:attSumScore}) to have a zero mean and unit standard deviation makes the scores more comparable.
This improves the accuracy of both the baseline (ENS) and our model (\hap{}) with the latter surpassing the former. 
However, this requires that the test data is available as a batch at the test time. 

The relative improvement in accuracy of our \hap{} model over the baseline is \mrkn{6}\% (aPaY), \mrkn{18}\% (AwA) and \mrkn{31}\% (CUB).
This trend nicely follows the level of granularity of the objects in the data sets.
Our model takes advantage of the underlying structure and the commonality among the classes and it is able to distinguish between fine grained classes more efficiently.
On the other hand, it is harder for the baseline models (DAP \& ENS) to distinguish such fine grained objects using the abstract \genatt attributes. 

In the rest of the experiments, we report the results when using the GoogLeNet features for our model and the baselines.

\makeatletter{}\begin{figure}[!t]
\centering
\begin{subfigure}[b]{0.5\linewidth}
	\centering
	\includegraphics[width=\linewidth]{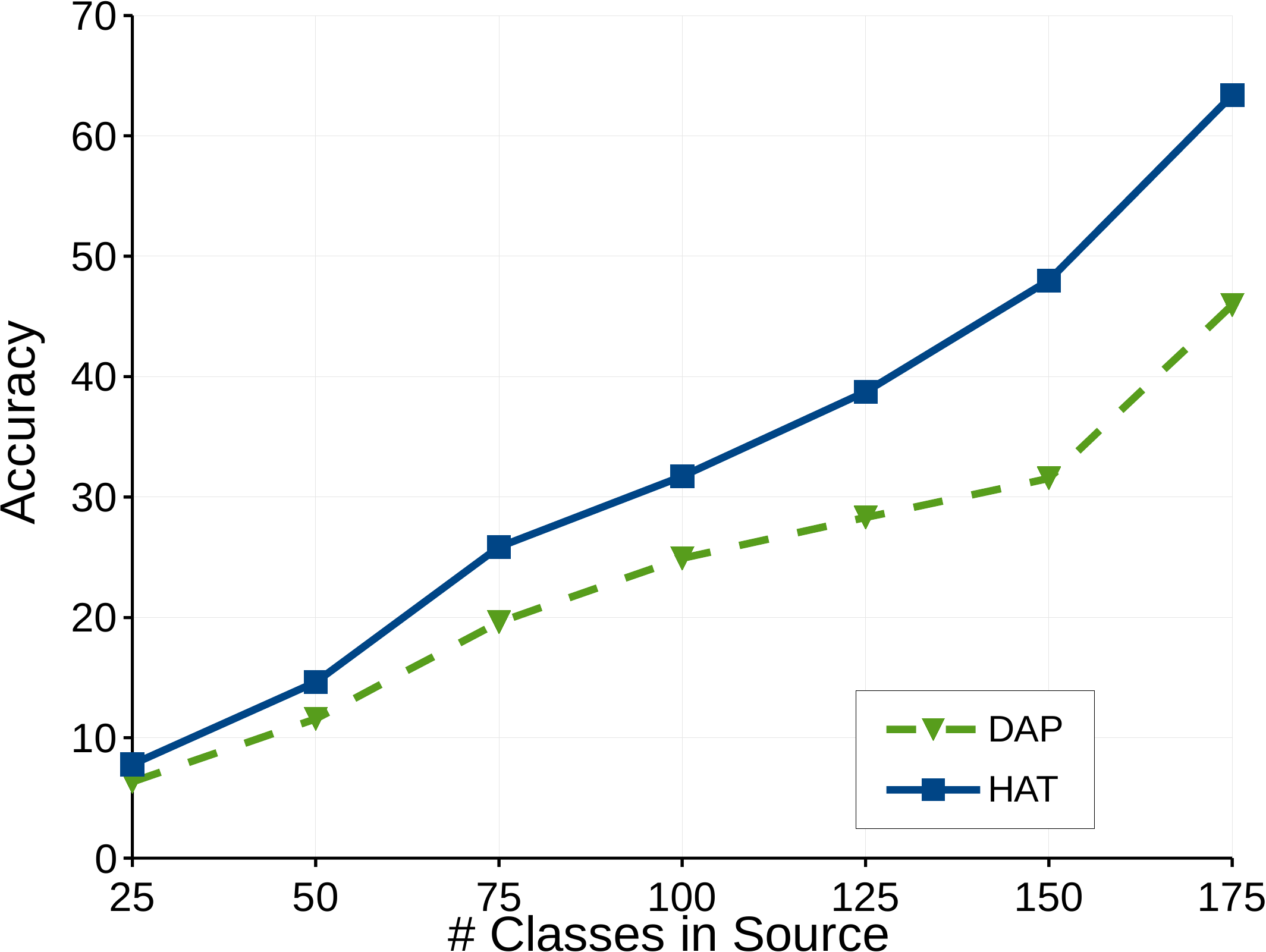}
	\caption{ }\label{fig:cub_acc}
\end{subfigure}~~
\begin{subfigure}[b]{0.5\linewidth}
	\centering
	\includegraphics[width=\linewidth]{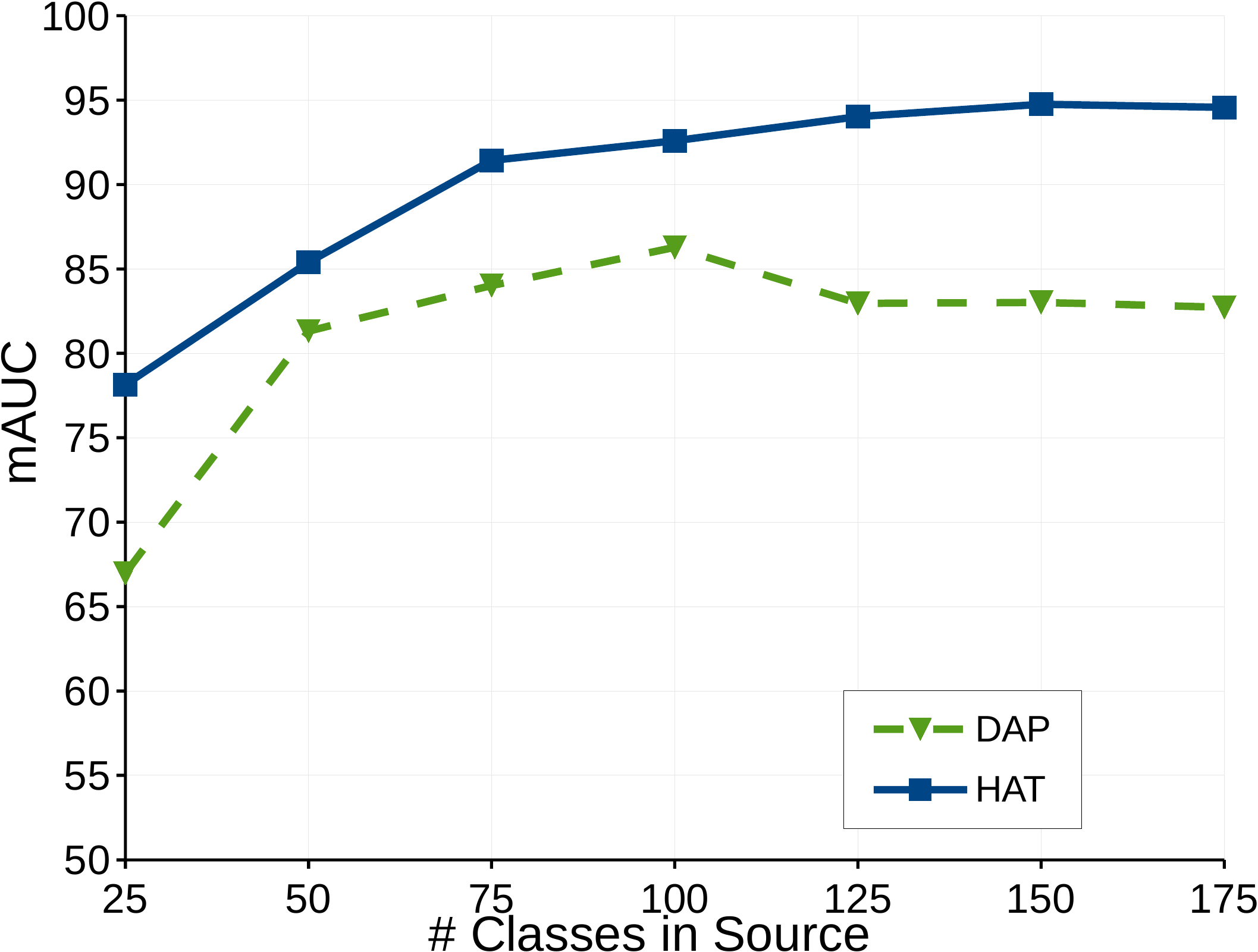}
	\caption{ }\label{fig:cub_auc}
\end{subfigure}~~
\figlblvspace
\caption{\mrkn{The performance of DAP and \hap{} in CUB with varying number of classes in the source as demonstrated with (a) multi-class accuracy and (b) mean AUC. (Features: GoogLeNet)} }
\label{fig:cub_source}
\figvspace
\end{figure} 
\textbf{Per-image vs. per-class attributes.} 
aPaY and CUB provide attribute annotation at the image level which we used to train the models in the previous experiments. 
We evaluate on these two datasets using class-based attributes similar to those in AwA. 
We notice that the accuracy of both the baseline and \hap{} decreases in these settings. 
Where ENS has \mrkn{38.3}\% and \mrkn{34.6}\%, the \hap{} model achieves \mrkn{40.3}\% and \mrkn{48.2}\% on aPaY and CUB. This seems to differ from the findings in \cite{Lampert2013}. 
The reason could be related to the type of features used. 
In \cite{Lampert2013} a set of shallow features are used which require a relatively larger number of samples to train good attribute classifiers.  
This in turn results in noisy attribute predictions if there are few image-based annotations of the attribute. 
In comparison, using the deep features we can learn better attribute classifiers even if the training data is relatively sparse.  

\makeatletter{}
 
\textbf{Unknown attributes of the novel class.} 
\makeatletter{}Although this evaluation setup is not possible with the \genatt attribute model, \hap{} enables us to carry out zero-shot recognition even if the attribute description of the novel class is unknown. 
To do that, we again leverage the hierarchy and transfer the attribute description of the parent node to the novel class. 
Using this setup, \hap{} achieves an accuracy of \mrkn{31.1}\% (aPaY), \mrkn{59.7}\% (AwA) and \mrkn{32.6}\% (CUB). 
This drop in performance is reasonable since we are transferring the more generic attributes of the parent. 
Hence, confusion can arise when multiple test classes share the same parent in the hierarchy. 
Nonetheless, \hap{} makes it possible to perform attribute-based zero-shot classification when only the label of the novel class is available.  

\textbf{Source set complexity.} 
\makeatletter{}In the following experiment, we vary the complexity of the knowledge contained in the source (the number of seen classes) compared to the target (the unseen classes).  
This helps to have a better understanding of the characteristics of the different models as the richness of the embedded information in the source changes. 
We use the CUB dataset and start with a random set of 25 classes to be in the source. 
We gradually increase the source set with additional 25 random classes. 
At each step, the rest of the 200 classes is used as the target set to conduct zero-shot classification. 

In \reffig{fig:cub_source} we see that when the source is relatively poor and contains less structured knowledge, both DAP and \hap{} performs at the same level. 
However, as the source get bigger and more complex \hap{} consistently outperforms DAP with an increasingly wider margin. 
Unlike DAP that uses a single layer of \genatt{} attributes, \hap{} is able to take advantage of the complexity of information available in the source. 
\hap{} captures the commonality among the categories and exploits it to learn and transfer more discriminative attributes to distinguish the unseen categories.  
 
\section{Conclusion}\label{sec:conclusion}
\makeatletter{}In this paper, we present a simple yet very effective model for zero-shot object recognition. 
Our model takes advantage of the embedded structure in the category space to learn attributes at different levels of abstraction. 
Furthermore, it exploits inter-class relations to provide a guided knowledge transfer approach that can select and transfer the expected relevant attributes to a novel class. 
The evaluation on three challenging datasets shows the superior performance of the proposed model over the state-of-the-art. 

We are considering to extend our approach in different directions. 
In the current model, we assume similar importance for the transferred attributes from the different layers in the hierarchy. 
However, the confidence in the attributes predictions and their relative relatedness to the novel class differ through the different abstraction levels. 
Using adaptive weights when transferring attributes could improve the model performance. 
Moreover, we considered only the ancestor nodes as a knowledge source. 
The siblings of a novel class represent another promising option. 
Including them in the transfer process may help in sharing more discriminative attributes.

\balance
{\reffontsize
\bibliographystyle{ieee}
\bibliography{arxiv16_ref}
}
\end{document}